\documentclass[10pt, a4paper]{article}

\usepackage[final]{lrec2026}

\usepackage{booktabs}
\usepackage{times}
\usepackage{latexsym}
\usepackage{subcaption}
\usepackage{rotating}
\usepackage{pdflscape}
\usepackage{caption}
\usepackage{subcaption} 
\usepackage{adjustbox}
\usepackage{hyperref}
\usepackage{makecell} 
\usepackage{tablefootnote}
\usepackage[T1]{fontenc}
\usepackage[utf8]{inputenc}
\usepackage{graphicx} 
\usepackage{multirow}
\usepackage{microtype}
\usepackage{inconsolata}
\usepackage{amsmath}
\usepackage{courier}
\usepackage{capt-of}
\usepackage{float} 
\usepackage[dvipsnames]{xcolor}

\title{Benchmarking NLP-supported Language Sample Analysis for Swiss Children’s Speech}

\name{Anja Ryser, Yingqiang Gao, Sarah Ebling*} 

\address{Department of Computational Linguistics \\
         University of Zurich, Switzerland \\
         \{ryser, yingqiang.gao, ebling\}@cl.uzh.ch\\}

\abstract{
Language sample analysis (LSA) is a process that complements standardized psychometric tests for diagnosing, for example, developmental language disorder (DLD) in children. 
However, its labour-intensive nature has limited its use in speech-language pathology practice. 
We introduce an approach that leverages natural language processing (NLP) methods that do not rely on commercial large language models (LLMs) applied to transcribed speech data from 119 children in the German-speaking part of Switzerland with typical and atypical language development. 
This preliminary study aims to identify optimal practices that support speech-language pathologists in diagnosing DLD more efficiently with active involvement of human specialists. Preliminary findings underscore the potential of integrating locally deployed NLP methods into the process of semi-automatic LSA.
 \\ \newline . }

\begin{document}
\pagestyle{empty}

\maketitleabstract



\makeatletter
\maketitleabstract
\begingroup
\renewcommand\thefootnote{}
\renewcommand\@makefnmark{}
\long\def\@makefntext#1{%
  \noindent\parbox{\dimexpr\columnwidth\relax}{#1}%
}
\footnotetext{%
*Corresponding author.
\hspace{1em}%
\includegraphics[height=0.32cm]{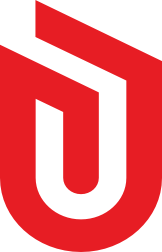}\, Dataset on 
\href{https://doi.org/10.48656/rqf2-sq76}{SWISSUBase}
}
\endgroup
\makeatother

\section{Introduction}

Developmental language disorder (DLD) is a neurodevelopmental condition, commonly diagnosed in children, that significantly affects an individual's ability to acquire and use spoken and written language, despite typically developed intelligence and no obvious sensory or neurological impairments or inadequate language exposure \citep{tomblin1996system, bishop2006causes, luke2023definition, van2024genetic}. 

As a recommended part of DLD diagnosis, language sample analysis (LSA) aims at evaluating the spontaneous\footnote{We acknowledge that elicited language is never completely ``spontaneous''; nevertheless, the term is common in connection with LSA.} language production skills of children \citep{gallagher2020measure, ramos2024using}. It involves collecting and analysing samples of language during conversation, storytelling, play, or other activities. LSA provides detailed information about a person's linguistic abilities, including vocabulary, grammar, sentence structure, and pragmatic language use. These insights can then be used in diagnostics, setting therapeutic goals and monitoring progress.

Despite being an effective tool for practice, LSA is not often used by speech-language pathologists due to its time- and effort-intensive process \citep{klatte2022language, bawayan2022language}. Modern NLP methods and machine learning can help to alleviate some of these challenges with their time efficient approaches to big amounts of data

\begin{table}[htb]
\centering
\small
\resizebox{\columnwidth}{!}{
\begin{tabular}{llcccc}
\toprule
\textbf{Transcription} & \textbf{Variant} & \textbf{WER} & \textbf{CER} & \textbf{MER} & \textbf{WIL} \\
\hline
\multirow{2}{*}{Original} 
& Swiss German       & 81.0 & 80.0 & 49.9 & 94.8 \\
& Swiss Std.\ German & 48.7 & 47.8 & 36.1 & 59.5 \\
\midrule
\multirow{2}{*}{Normalized} 
& Swiss German       & 57.2 & 55.7 & 38.6 & 73.8 \\
& Swiss Std.\ German & 45.6 & 44.8 & 35.2 & 54.7 \\
\bottomrule
\end{tabular}
}
\caption{Average ASR results on Whisper transcriptions for original and normalized text.}
\label{tab:asr_results}
\end{table}

\begin{figure*}[htb]
    \centering
    \includegraphics[width=\textwidth]{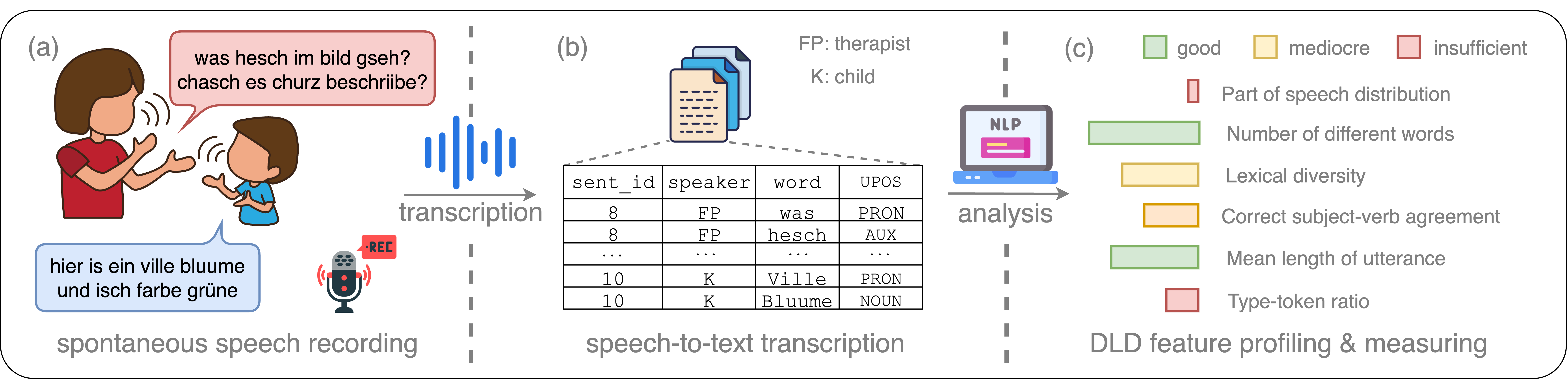}
    \caption{Our pipeline of LSA with NLP-supported approaches for diagnosis of DLD. \textbf{(a) Spontaneous speech recording}: a speech-language pathologist interacts with the child in a naturalistic setting and both of their speeches are recorded; \textbf{(b) Speech-to-text transcription}: the recordings are (automatically) converted into text and post-corrected by the speech-language pathologist and further (automatically) tokenized to words; \textbf{(c) DLD feature profiling and measuring}: approaches such as POS tagging, dependency parsing, stemming and lemmatization, etc., are applied to create the DLD feature profiles, where various linguistic measures are computed to evaluate the language abilities of children. The final diagnostic decision is made by the human expert (i.e., the speech-language pathologist), taking into account the output of the pipeline as well as other criteria. The speech utterances demonstrated are in Swiss German.}
\end{figure*}

In this study, we evaluate the zero-shot capability of non-commercial NLP methods, namely ASR and part-of-speech (POS) tagging on data collected from 110 children living in Switzerland. These NLP methods are the foundation on which analyses of the speech and language samples, such as measurements for lexical density and diversity and many others, are based on. To ensure the quality and reliability of these analyses, it is crucial to have trustworthy NLP methods with no commercial LLMs involved, as described in this study.

With data collected from 119 children living in Switzerland, we present a case study demonstrating that NLP approaches that do not rely on commercial LLMs can effectively assist speech-language pathologists in identifying critical linguistic patterns essential for LSA. Results from analyses of Swiss German and Swiss Standard German speech transcriptions highlight the potential of automating LSA. To achieve the high quality performance needed in a clinical setting, our results also show the need for specific training and fine-tuning.

Our \textbf{main contributions} are three-fold: 1) We demonstrate a possible annotation process in semi-automated LSA for the elicitation of clinical linguistic features of DLD in Swiss German; 2) We release the first case study dataset in Swiss German and Swiss Standard German containing six speech transcriptions together with their annotations; 3) We provide an assessment of the effectiveness of NLP methods not based on commercial LLMs for LSA on both Swiss German and Swiss Standard German.

\section{Related Work}

\subsection{Speech Transcription}

Transcribing children's speech into text is a critical first step in LSA. However, advanced end-to-end approaches, such as ASR, still face substantial challenges due to the high inter- and intra-speaker variability in pronunciation, vocabulary, and speech rate among children across different ages as well as higher fundamental frequencies in children \citep{potamianos1997automatic, bhardwaj2022automatic}. 
\citet{smith2017improving} trained a deep neural network on out-of-domain adult speech data, which was subsequently fine-tuned using speech data from children with DLD. Similarly, \citet{rumberg2021age} proposed a framework for age-invariant training, leveraging age-independent patterns derived from both adult and child speech. \citet{jain2023adaptation} studied the adaption of the Whisper model \citep{radford2023robust} to child speech via self-supervised fine-tuning. \citet{pokel2025diss} introduced a novel algorithm for ASR tailored to dysarthric German speech, which restructures word-level utterances into sentence-level sequences. This approach demonstrates promising results in improving the accessibility of speech transcriptions. 

These approaches, often leveraging the power of transfer learning by utilizing adult speech data, have shown effectiveness in increasing performance on children's speech data. However, the fundamental lack of data collected from children with DLD still limits the applicability of ASR in LSA. 
In recent years, ASR systems for low-resource languages, such as Swiss German, have been developed \citep{kew2020uzh, nigmatulina2020asr, arabskyy2021dialectal, schraner2022swiss, timmel2024finetuningWhisper}, supported by the availability of corpora of Swiss German speech data \citep{pluss2020swiss, dogan2021swissdial, pluss2022sds, pluss2023stt4sg, stucki2025swissgpc}. However, due to licensing limitations, these models are not currently available and thus the ability of Swiss German models to generalize to speech data of children with DLD remains largely unexplored. Additionally, finetuning of commercial models running on commercial computation services with children's speech data is heavily restricted due to legal and ethical considerations and limitations. \citep{liu2024automatic}. As Whisper models perform well on Swiss German in a zero-shot setting \citep{dolev2024does} and their availability, a Whisper model was used for this research for speech transcription.

\subsection{Feature Analysis}
\label{section: feature_analysis}

Initial efforts to develop semi-automated LSA approaches have emerged in the past years, with a primary focus on English speech data.
\citet{gabani2011exploring} analysed speech data collected from monolingual English-speaking children and proposed NLP methods to predict the presence of DLD. The study utilized eight categories of linguistic features (later expanded by \citet{hassanali2012evaluating} to include syntactic and semantic features) to train language models as predictors, which provided important aspects for future research. 
\citet{solorio2013survey} provided a concise summary of the types of NLP-features employed in LSA for the diagnosis of DLD and highlighted questions for future research.
\citet{ludtke2023multidisciplinary} described the ideal hypothetical system capable of recording spontaneous speech of children while effectively separating background noise and speech from non-target individuals. This system can transcribe and segment recorded speech, offering a wide range of measurements, including environmental factors of recording, DLD profiles, and detailed analyses across various linguistic structures and elements.

These studies have laid the groundwork for advancing semi-automated LSA approaches, highlighting key linguistic features, methodological considerations, and future directions for improving DLD diagnosis. However, none of these works investigated LSA for children's speech in Swiss German.

DLD features can manifest in all linguistic categories. At the moment we are focusing on the grammatical level, which can be analysed based on transcripts of speech. As part of future research, phonetic and phonological features could also be analysed directly on the speech recordings.

In this work, we show that it is possible to perform LSA using NLP methods that are not based on commercial LLMs and are both ethical and effective. While we acknowledge LLMs are likely to dominate in LSA, we argue that other NLP methods still deliver good results without causing potential ethical issues.

\section{Data}

The dataset is scheduled to be published on SWISSUbase (DOI: \url{https://doi.org/10.48656/rqf2-sq76}) and will be released in the near future for public access.

\subsection{Data Collection}

Speech data collected from children are highly sensitive and require careful handling. In compliance with the regulations of the responsible research committee, we obtained the necessary ethical approval to collect speech utterances, accompanied by signed consent forms from the children's parents.\footnote{For further details, please refer to the Ethics Statement section.} Speech utterances were recorded in both therapeutic and naturalistic settings (i.e., kindergarten), capturing spontaneous interactions between one therapist and one child. To ensure privacy, the data were stored in an anonymized format, with no metadata linked to identifiable codes. 

\begin{table}[htb]
    \centering
    \resizebox{\columnwidth}{!}{
        \begin{tabular}{lrr}
    \toprule
    & \textbf{Swiss German} & \textbf{Swiss Std. German} \\
    \hline
    \# recordings & 91 & 19\\ 
    \# hours & 17:57 & 3:35 \\
    \# utterances & 16,553 & 3,014\\ 
    \# words & 126,733 & 21,356\\

    \bottomrule
    \end{tabular}
    }
    \caption{Statistics of the collected data. The table summarizes the total duration of recordings, the number of utterances and individual words for both speakers.}
    \label{tab:statistics}
\end{table}

We collected speech samples of duration between 10 and 20 minutes from 119 children with typical and atypical Swiss German and Swiss Standard German speech and 25 speech-language pathologists. Of these, we obtained permission to publish the data of 41 recordings. Although the dataset is relatively small, our objective is to initiate research into the application of semi-automated LSA for Swiss German speech utterances based on this dataset. The recordings were made with phones in standard quality and stored as mp3-files.

\subsection{Data Statistics}

Table~\ref{tab:statistics} presents the overall statistics of the collected speech data. Speech recordings were obtained in both Swiss German and Swiss Standard German. The ages of the children range from four to eight, encompassing an important phase in the assessment of and intervention for DLD \citep{sansavini2021developmental}. 
The speech data were transcribed by students of speech and language therapy and subsequently verified by professionals native in both Swiss German and Swiss Standard German. 

We engaged with children living across Switzerland. Therefore, the collected recordings are composed of different Swiss German dialects such as ``\textit{Z\"urid\"u\"utsch}'' (dialect spoken in the Canton of Zurich) and ``\textit{Baseld\"u\"utsch}'' (dialect spoken in the Canton of Basel).

\section{Methods}

Starting from the collected recordings of spontaneous speech of Swiss children, we showcase our data processing methods specifically used for semi-automated LSA. 

\subsection{Speech Transcription}
The first task is to transcribe speech into text based on the raw audio recordings. To do this, we apply two approaches:
\paragraph{Manual transcription by human experts.}
We recruited 13 students majoring in speech and language therapy, who had received training in the transcription method used for this study. For the transcriptions we use an adapted form of the \textit{Dieth-Schreibung} (\textit{Dieth spelling}) \citep{dieth1986schwyzertütsch} with added annotation of features important for language and speech pathology, such as annotation of stress, unintelligible sequences, pauses, mazes, and overlaps in turn taking. Each transcript was created by one student and checked by another. The manual transcriptions served as the ground-truth reference for our study. See Table~\ref{tab:annotation-examples} in Appendix~\ref{appendix:annotation-examples} for dataset examples. Additionally, the authors normalized the transcript manually to compare the influence on non-standard orthography and an assimilation to Standard German. Where possible, orthographically correct versions of words were used while keeping the word order intact.

\paragraph{Transcription using ASR models.}
We deployed a Whisper model (\citet{radford2023robust}, Hugging Face checkpoint \texttt{openai/whisper-small}\footnote{Model available at \url{https://huggingface.co/openai/whisper-small}}) \textbf{locally} to transcribe speech recordings into spoken sentences. This lightweight Whisper model, with 244 million parameters, was selected for its ease of deployment on local computers to prevent data leakage and its multilingual transcription capabilities and before the data was collected. 

Automatically transcribing the speech samples used in this study posed three challenges: (1) Most large ASR systems are not trained with Swiss German data; (2) Most ASR systems underperform in transcribing speech from children \citep{bhardwaj2022automatic}; (3) The speech transcribed contains non-standard, atypical or wrong grammar due to the atypical language development of the children. Combining these three difficulties is challenging, requesting highly specifically trained models to solve the task reliably.

Manual transcription of spontaneous speech demands substantial knowledge and expertise in LSA and considerable time investment. Therefore, we aimed to evaluate the performance of the state-of-the-art Whisper model on the speech transcription task, given its potential to significantly reduce the workload of speech-language pathologists in practice.

\subsection{Part-of-speech (POS) Tagging}
\label{section:pos-tagging}

Since POS tagging delivers information for certain linguistic features used for the feature analysis as described in Section \ref{section: feature_analysis}, and thus helps in identifying morphosyntactic errors in language samples, our second task is to perform POS tagging for transcriptions in Swiss German and Swiss Standard German. For an overview over both POS tag sets used in this work, see Table \ref{tab:pos-tagging}.

\begin{table}[h!]
    \centering

    \begin{subtable}[t]{\columnwidth}
    \resizebox{\columnwidth}{!}{
    \begin{tabular}{llll}
    \toprule
     \textbf{UPOS Tags} & \textbf{Feature} & \textbf{UPOS Tags} & \textbf{Feature}  \\
     \midrule
     ADJ & adjective & ADP & adposition \\ 
     ADV & adverb & AUX & auxiliary \\
     CCONJ & co. conjunction & DET & determiner \\
     INTJ & interjection & NOUN & noun \\
     NUM & numeral & PART & particle \\
     PRON & pronoun & PROPN & proper noun \\
     PUNCT & punctuation & SCONJ & sub. conjunction \\
     SYM & symbol & VERB & verb \\
     X & other \\ 
    \end{tabular}
    }
    \end{subtable}
    \vspace{1em}
    \begin{subtable}[t]{\columnwidth}
    \resizebox{\columnwidth}{!}{
    \begin{tabular}{ll}
    \midrule
     \textbf{STTS Tags} & \textbf{Feature} \\
     \midrule
     ADJA & Attributive adjectives \\
     ADJD & Predicative or adverbial adjectives  \\
     APPO & Postpositions \\
     APPR & Prepositions \\
     APPRART & Prepositions with an article \\
     APZR & Circumpositions (right part) \\
     ADV & True adverbs \\
     ART & Definite/indefinite articles \\
     CARD & Cardinal numbers \\
     KOKOM & Comparative particles \\
     KON & Coordinating conjunctions \\
     KOUI & Subordinating conjunctions with infinitive \\
     KOUS & Subordinating conjunctions with clauses \\
     ITJ & Interjections \\
     NE & Proper nouns \\
     NN & Common nouns \\
     PTKA & Particles with adjectives or adverbs \\
     PTKANT & Response particles \\
     PTKNEG & Negation particles \\
     PTKVZ & Separable verb prefixes \\
     PTKZU & ``zu'' before infinitives \\
     PAV & Pronominal adverbs \\
     PDAT & Attributive demonstrative pronouns \\
     PDS & Substituting demonstrative pronouns \\
     PIAT & Attributive indefinite pronouns without determiners \\
     PIDAT & Attributive indefinite pronouns with determiners \\
     PIS & Substituting indefinite pronouns \\
     PPER & Non-reflexive personal pronouns \\
     PPOSAT & Attributive possessive pronouns \\
     PPOSS & Substituting possessive pronouns \\
     PRF & Reflexive personal pronouns \\
     PRELAT & Attributive relative pronouns \\
     PRELS &  Substituting relative pronouns \\
     PWAT & Attributive interrogative pronouns \\
     PWAV & Adverbial interrogative pronouns \\
     PWS & Substituting interrogative pronouns \\
     TRUNC & Truncation \\
     FM & Foreign language material \\
     XY & Non-words \\
     VAFIN & Auxiliary finite verbs \\
     VMFIN & Modal finite verbs \\
     VVFIN & Full finite verbs \\
     VAIMP & Auxiliary imperative verbs \\
     VVIMP & Full imperative verbs \\
     VAINF & Auxiliary infinitives \\
     VMINF & Modal infinitives (substitute infinitive) \\
     VVINF & Full infinitives \\
     VVIZU & Infinitives with ``zu'' \\
     VAPP & Auxiliary past participles \\
     VMPP & Modal past participles \\
     VVPP & Non-inflected full past participles \\
    \bottomrule
    \end{tabular}
    }   
    \end{subtable}
    \caption{Overview of two POS tagging systems for German, UPOS (top) and STTS (bottom). The core difference is that STTS contains the level of detail required for LSA.}
    \label{tab:pos-tagging}
\end{table}

For Swiss German transcriptions, we applied two BERT-based models \citep{aepli2022improving}, trained with two different POS tag sets: 1) \texttt{swiss\_german\_pos\_model}, trained with the Universal POS tags (UPOS\footnote{\url{https://universaldependencies.org/u/pos/}}), and 2) \texttt{swiss\_german\_stts\_pos\_model}, trained with the Stuttgart-Tübingen-Tagset (STTS\footnote{More details available at \url{https://homepage.ruhr-uni-bochum.de/stephen.berman/Korpuslinguistik/Tagsets-STTS.html}}). Both models were deployed \textbf{locally} to adhere to the data privacy rules. 

For Swiss Standard German transcriptions, we tested the statistical model \texttt{de\_core\_news\_sm} provided by \texttt{spaCy}\footnote{\url{https://spacy.io/models/de}, MIT licence} as a supplement to the BERT-based models.

The models were chosen due to their availability before the data was collected.

\paragraph{Inter-annotator Agreement}
We recruited three native Swiss German speakers and three speakers with profound knowledge of Swiss Standard German with strong background in computational linguistics to annotate the gold-standard UPOS and STTS tags for sentences in the manual speech transcriptions. These annotations serve as ground truth for evaluating the performance of the three models discussed in Section~\ref{section:pos-tagging}. Prior to initiating the annotation process, we conducted training sessions with all annotators, during which the task instructions were explained in detail. We attached the instruction sheet in Appendix~\ref{section:instruction} for reference.

\begin{table}[htb]
    \centering
    \begin{subtable}[t]{0.5\textwidth}
        \centering
    \resizebox{0.95\columnwidth}{!}{
        \begin{tabular}{lcc}
            \toprule
            & \textbf{Swiss German} & \textbf{Swiss Std. German} \\
            \midrule
            A\&B & 0.804 & 0.861 \\ 
            B\&C & 0.850 & 0.844 \\ 
            A\&C & 0.802 & 0.900 \\
            \bottomrule
        \end{tabular}
        }
        
        \caption{IAA for UPOS tagging.}
        \label{tab:IAA-UPOS}
    \end{subtable}
    \hfill
    \begin{subtable}[t]{0.5\textwidth}
        \centering
        \resizebox{0.95\columnwidth}{!}{
        \begin{tabular}{lcc}
            \toprule
            & \textbf{Swiss German} & \textbf{Swiss Std. German} \\
            \midrule
            A\&B & 0.910 & 0.926 \\ 
            B\&C & 0.939 & 0.921 \\ 
            A\&C & 0.926 & 0.945 \\
            \bottomrule
        \end{tabular}
        }
        
        \caption{IAA for STTS tagging.}
        \label{tab:IAA-STTS}
    \end{subtable}
    \caption{Pairwise linearly weighted Cohen's Kappa \citep{cohen1968weighted} among three human annotators who are native Swiss German speakers for UPOS and STTS tagging on Swiss German (100 sentences) and Swiss Standard (Std.) German (80 sentences) transcriptions. }
    \label{tab:IAA-main}
\end{table}

\begin{table}[htb]
    \centering
    \label{tab:linguistic_features}
    \resizebox{0.8\columnwidth}{!}{
    \begin{tabular}{ll}
        \toprule
        \textbf{Key} & \textbf{Value} \\ 
        \midrule
        Case & Acc, Nom, Gen, Dat \\ 
        Number & Sing, Plur \\ 
        Gender & Fem, Masc, Neut \\ 
        Person & 1, 2, 3 \\ 
        PronType & Art, Dem, Ind, Int, Prs, Rel \\ 
        Mood & Ind, Sub, Imp \\ 
        Tense & Past, Pres \\ 
        VerbForm & Fin, Inf, Part \\ 
        Definite & Def, Ind \\ 
        Degree & Cmp, Pos, Sup \\ 
        Foreign & Yes \\ 
        Poss & Yes \\ 
        Reflex & Yes \\ 
        \bottomrule
    \end{tabular}
    }
    \caption{Values for morphological categories. Notice that morphological categories rely on the \texttt{spaCy} model \texttt{de\_core\_news\_sm} and are therefore language- and model-dependent.}
    \label{tab:linguistic_values}
\end{table}

\begin{table*}[htb]
    \centering
    \begin{adjustbox}{max width=0.9\textwidth}
    \begin{tabular}{ll}
        \toprule
        \textbf{Tag} & \textbf{Key-Value Features} \\
        \midrule
        ADJA & \{`Case': `Acc', `Degree': `Pos', `Gender': `Fem', `Number': `Plur'\} \\
        ADJD & \{`Degree': `Pos'\} \\
        APPRART & \{`Case': `Dat', `Gender': `Neut', `Number': `Sing'\} \\
        ART & \{`Case': `Nom', `Definite': `Def', `Gender': `Neut', `Number': `Sing', `PronType': `Art'\} \\
        NN/NE & \{`Case': `Acc', `Gender': `Masc', `Number': `Sing'\} \\
        PDS/PDAT (plural) & \{`Case': `Nom', `Number': `Plur', `PronType': `Dem'\} \\
        PDS/PDAT (singular) & \{`Case': `Nom', `Gender': `Fem', `Number': `Sing', `PronType': `Dem'\} \\
        PIS & \{`Gender': `Neut', `PronType': `Ind'\} \\
        PIAT & \{`Case': `Nom', `Gender': `Fem', `Number': `Sing', `PronType': `Ind'\} \\
        PPER (other)  & \{`Case': `Nom', `Number': `Sing', `Person': `3', `PronType': `Prs'\} \\
        PPER (singular) & \{`Case': `Nom', `Gender': `Neut', `Number': `Sing', `Person': `3', `PronType': `Prs'\} \\
        PPOSAT & \{`Case': `Acc', `Gender': `Fem', `Number': `Sing', `Poss': `Yes', `PronType': `Prs'\} \\
        PPOSS & \{`Case': `Nom', `Gender': `Masc', `Number': `Sing', `Poss': `Yes', `PronType': `Prs'\} \\
        PRF & \{`Case': `Acc', `Number': `Sing', `Person': `3', `PronType': `Prs', `Reflex': `Yes'\} \\
        PRELS/PRELAT & \{`Case': `Acc', `Gender': `Neut', `Number': `Sing', `PronType': `Rel'\} \\
        PWS & \{`Case': `Nom', `Gender': `Masc', `Number': `Sing', `PronType': `Int'\} \\
        PWAV & \{`PronType': `Int'\} \\
        VERB & \{`Mood': `Ind', `Number': `Sing', `Person': `3', `Tense': `Pres', `VerbForm': `Fin'\} \\
        VERB IMP & \{`Number': `Sing'\} \\
        VERB INF & \{`VerbForm': `Inf'\} \\
        VVPP & \{`VerbForm': `Part'\} \\
        \bottomrule
    \end{tabular}
    \end{adjustbox}
    
    \caption{Morphological features for STTS tags (based on \texttt{spaCy}), sorted alphabetically. The UPOS tags are converted to STTS tags using a conversion look-up table.}
    \label{tab:linguistic_tags}
\end{table*}

\subsection{Morphological Features}
German is a morphologically rich language that utilizes all 17 UPOS tags. For instances where a single POS tag is insufficient to capture lexical distinctions — such as the straightforward example of nouns in German, which can exhibit three different genders: masculine, feminine, and neutral—, features that describe these linguistic differences become essential. 

To broaden the scope of our study, we annotated the morphology of language samples by assigning specific values to identified linguistic features (i.e., keys in our annotations). 
Table \ref{tab:linguistic_values} provides a summary of the morphological values associated with these keys. 
Additionally, Table \ref{tab:linguistic_tags} outlines the mappings between POS tags and linguistic features, presented as key-value pairs. 

The automatic prediction of morphological key-value features can be achieved using traditional statistical models \citep{can2011statistical, silfverberg2011combining} or deep learning approaches \citep{tkachenko2018modeling, bohnet2018morphosyntactic, klemen2023enhancing}, both of which require a substantial amount of Swiss German language samples. However, deep learning approaches require a large amount of training data, which is currently absent in the Swiss LSA research.

To address this requirement, we are working on collecting additional language samples to support the training of these models. A comprehensive investigation of morphology prediction, however, is beyond the scope of this study and will be comprehensively addressed in future work.

\section{Results}

\subsection{ASR Transcriptions}

\begin{figure*}[htb]
    \centering
    \begin{subfigure}[t]{0.43\textwidth}
        \centering
        \includegraphics[width=\linewidth]{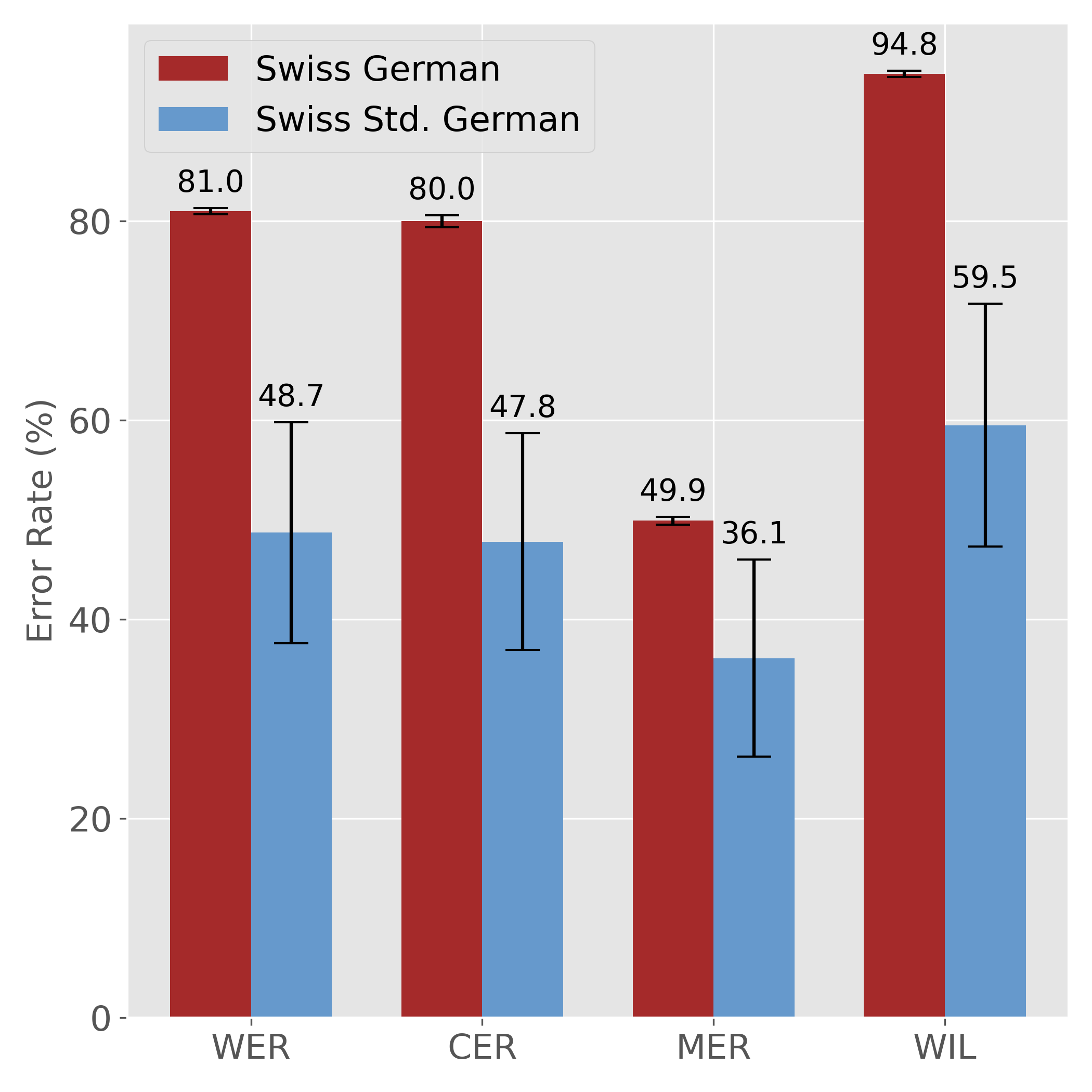}
        \caption{Original Transcriptions.}
        \label{fig:original-transcriptions}
    \end{subfigure}
    \hfill
    \begin{subfigure}[t]{0.43\textwidth}
        \centering
        \includegraphics[width=\linewidth]{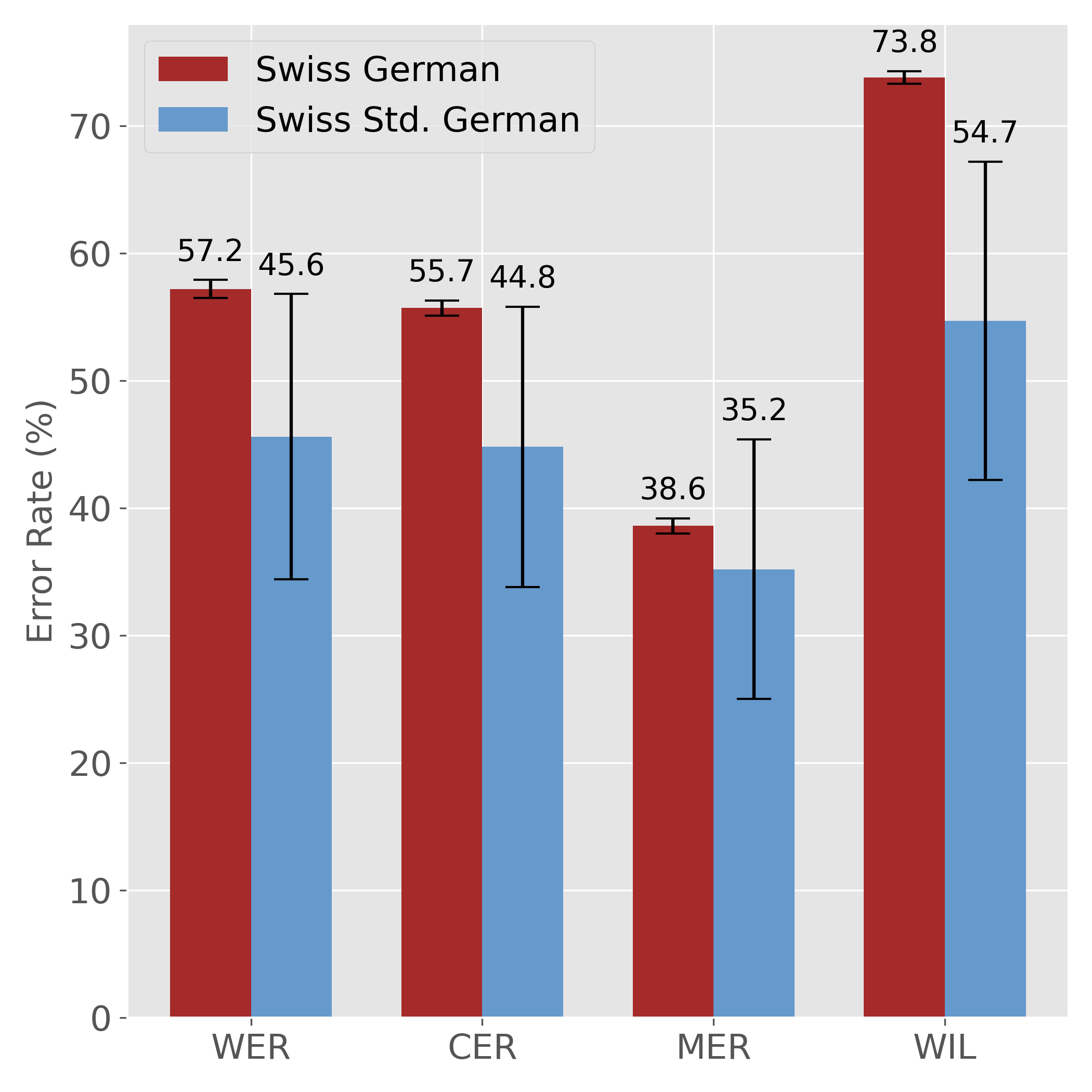}
        \caption{Normalized Transcriptions.}
        \label{fig:normalized-transcriptions}
    \end{subfigure}
    \caption{Average ASR results on Whisper transcriptions with standard deviations.}
    \label{fig:ASR-results}
\end{figure*}

\begin{figure*}[htb]
    \centering
    \includegraphics[width=\textwidth]{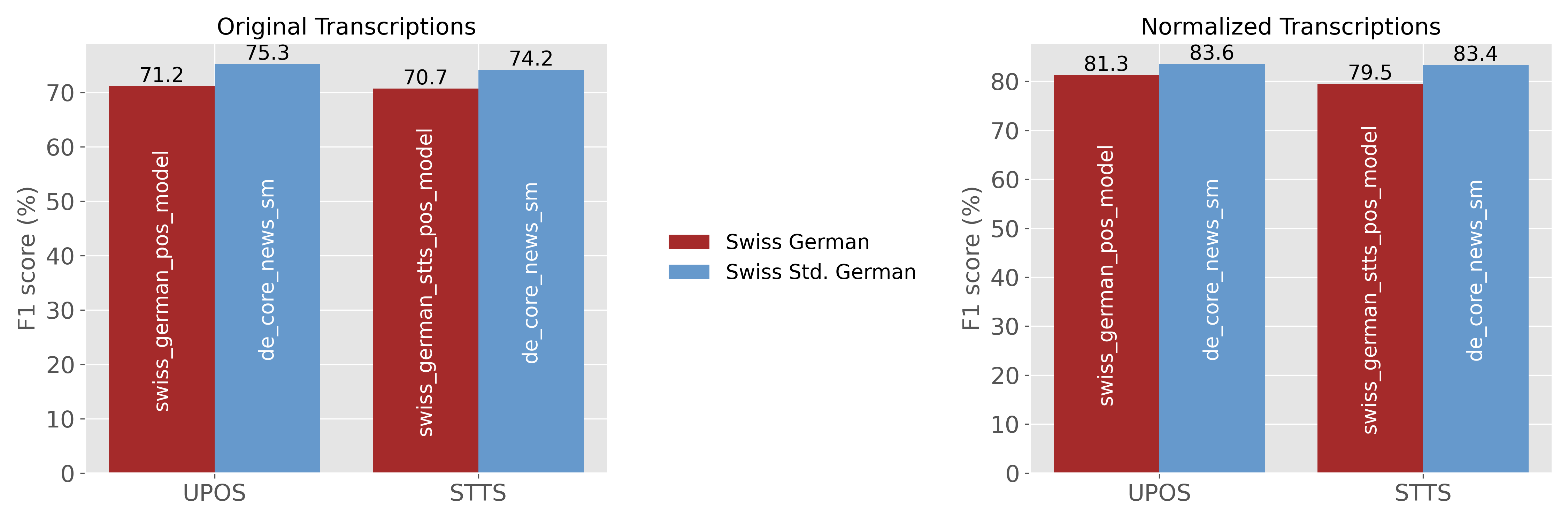}
    \caption{POS tagging results with BERT-based POS tagging model and \texttt{spaCy} model, measured on all transcription data for Swiss German and Swiss Standard German.}
    \label{fig:pos-tagging-results}
\end{figure*}

In Figure~\ref{fig:ASR-results}, we visualize the error rates of the Whisper-based ASR model transcribing the spontaneous speech recordings between one therapist and one child (Table~\ref{tab:asr_results}). We report Word Error Rate (WER), Character Error Rate (CER), Match Error Rate (MER), and Word Information Lost (WIL) of Swiss German and Swiss Standard German transcriptions.\footnote{We use the \href{https://github.com/jitsi/jiwer}{\texttt{JiWER}} Python package for all error rate calculations, Apache-2.0 license.} 
The WER for Swiss German is 81\%, and 48.7\% for Standard German. After normalizing the transcripts, the WER is 57.2\% for Swiss German and for Standard German 45.6\% 

\subsection{Inter-annotator Agreement on Human Labelling of POS Tags}

Inter-annotator agreement (IAA) scores were calculated based on 100 sentences for Swiss German and 80 sentences for Swiss Standard German, as one annotator for Swiss Standard German was unable to complete annotations for all 100 sentences. These sentences were randomly sampled from the corresponding transcriptions.

Overall, we achieved high IAA scores for both UPOS tagging (above 0.8) and STTS tagging (above 0.9), as measured using linearly weighted Cohen's Kappa \citep{cohen1968weighted}. The detailed scores are presented in Table~\ref{tab:IAA-main}. 

\subsection{Automatic POS Tagging}

In Figure~\ref{fig:pos-tagging-results}, we present the POS tagging performance of the BERT-based models and the \texttt{spaCy} model, evaluated on the complete transcription datasets for both Swiss German and Swiss Standard German. All models achieved F1 scores above 70, with significantly higher results observed on normalized transcriptions (above or nearly 80) compared to original transcriptions.

\section{Discussion}

\subsection{Challenges of Manual and Automatic LSA}
LSA poses many challenges despite being a gold standard tool in speech therapy \citep{klatte2022language, bawayan2022language}, which have to be taken into account in the manual use as well as in the development of a semi-automatic pipeline for LSA. Factoring them in from the first steps of development as well as searching for solutions is part of ongoing and future research.

\paragraph{Data collection.} Speech and language samples of children are typically recorded in naturalistic environments, such as in free conversations, which include both relevant speech and background noise like TV soundtracks \citep{ludtke2023multidisciplinary}, or, in our case, noise of other children playing and street noise in the background. Extracting meaningful speech utterances and transcribing them into text units which speech-language pathologists can directly work with, is often a challenging task. 

\paragraph{Manual annotation.} Linguistic features such as errors on the levels of syntax, morphology, semantics, or phonology are often manually annotated and analysed by speech-language pathologists, which due to its time-intensive nature makes the use in therapeutic assessment less common \citep{owens2018moving}.

\paragraph{Generalization difficulty.} Language error patterns (such as wrong word order) vary from one child to another and from dialect to dialect, which limits the generalizability of LSA methods.

\paragraph{Use of LLMs} In recent years, large language models (LLMs) have undergone significant advancements. However, querying commercial LLMs such as ChatGPT with research data, including language samples from children, is not compliant with Swiss data protection regulations. Furthermore, the use of commercial LLMs for language sample analysis (LSA) is neither ethical nor practical due to the highly sensitive nature of children’s speech transcriptions and the risk of data contamination. As a result, progress in automating LSA for the diagnosis of developmental language disorder (DLD) remains limited and understudied.

\paragraph{Swiss German} The linguistic landscape of Switzerland presents unique challenges for applying NLP methods \citep{parida2020idiap}. The prevalence of Swiss German dialects, which differ significantly from Standard German in their linguistic structure, complicates real-world NLP practice. Automatic speech recognition (ASR) on Swiss German typically produces Standard German transcriptions, thereby omitting or distorting important dialectal information or not being able to represent Swiss German faithfully, for example sentences such as ``Ich gang go poste.'' (\textit{I'm going shopping.}) can be translated into ``Ich gehe einkaufen.'', losing the ``go'', which does not exist in German but is essential in Swiss German. Models trained on written Standard German typically perform poorly on Swiss German due to the lack of a standardized written form \citep{kew2020uzh, nigmatulina2020asr}. Despite this, written Swiss German is increasingly used in digital communication, such as social media and online messaging, where individual writing styles introduce further variation \citep{hollenstein2014compilation}. 
In educational settings, children use Swiss Standard German. Swiss Standard German is a variety of Standard German but still contains words (``helvetisms'') and grammar rules that render it different from Standard German of Germany or Austria. Many children who do not speak Swiss German as their native language primarily, or exclusively, communicate in Swiss Standard German. This highlights the necessity of NLP methods that can effectively process both Swiss Standard German and Swiss German while accommodating the inherent variation within Swiss German dialects.

Research in other languages such as English supports the effectiveness of LSA in diagnosing DLD \citep{ramos2022sharpening} and the effectiveness of using automated LSA \citep{miller1985systematic, pye1994childes}. However, relevant studies are less present for German data and entirely missing for Swiss German. Our preliminary study investigates the potential of using NLP methods as a step closer to working with Swiss German data.

\subsection{ASR Transcriptions}
The majority of ASR transcription errors committed by the Whisper model during the experiments can be attributed to children speech. Specifically, Whisper models often failed to recognize the children's utterances (but not the utterances of the adult therapists) or generated repeated words. After orthographically correcting the manual transcriptions (i.e., normalization; for examples, see columns \textbf{word} (original) and \textbf{normalized} in Appendix~\ref{appendix:annotation-examples}), the error rates were significantly reduced (see Figure~\ref{fig:normalized-transcriptions}). This highlights the continued need for normalization of automatic speech transcriptions in practice to address the limitations of ASR models. This problem could also potentially be mitigated by fine-tuning the models with specific methods of transcription, which would allow keeping the important transcribed information as well as reaching a sufficient quality of transcription.

In our case study, indicated by the less prominent results of the Whisper speech transcriptions, fully relying on ASR outputs was not meaningful at this stage due to the limited availability of children’s speech data in Swiss German, which precluded fine-tuning even the small variant of the Whisper model. This highlights the importance of both naturally expanding the dataset and exploring alternative approaches, such as data augmentation via speech synthesis using generative text-to-speech models \citep{ren2021fastspeech, ao2022speecht5, toyin2024sttatts, hu2026qwen3}, to address the challenges of processing underrepresented and atypical speeches in a speech-language pathology context.

\subsection{Inter-annotator Agreement on Human Labelling of POS Tags}

Our results indicate a strong level of agreement, demonstrating the reliability of our annotation process. Prior research on IAA for German and Latin POS tagging \citep{brants2000inter, stussi2024part} has reported even higher IAA scores with professionally trained annotators, which suggests that the annotation quality could be further improved with expert training.

We argue that three primary factors contribute to the differences observed in our study: (1) While no major discrepancies exist, Swiss Standard German differs from the Standard German used in Germany, which can lead to disagreements among annotators; (2) Swiss German lacks standardized spelling, and our annotators originate from different Swiss German-speaking cantons, resulting in variations in their interpretation of syntactic functions; and (3) Due to the presence of erroneous, incomplete, or atypical sentence structures produced by children with DLD, some ambiguity remained, leading to variability in interpretation among annotators. In Appendix~\ref{section:pos-tag-error}, we provide examples of common discrepancies between our annotators.

\subsection{Automatic POS Tagging}
The results for the more general statistical \texttt{spaCy} model on Swiss Standard German data are consistently outperforming the two BERT-based POS tagging models specifically trained on UPOS and STTS for Swiss German annotations on the Swiss German data, further highlighting the difficulties specific to Swiss German in NLP methods. All in all, these findings highlight that the models used in this study can perform reasonably well on Swiss German and Swiss Standard German transcriptions while offering the advantages of being significantly less computationally expensive and more ethical in their deployment compared to commercial LLMs. More Swiss German training data in general, as well as finetuning the model to the specifics of our dataset (e.g. children's' speech and atypical speech) is expected to improve the performance.

\subsection{Current Challenges}
Despite achieving a reasonably good performance in automatic speech transcription and POS tagging, it is important to emphasize that the current results remain insufficient for therapeutic practice. This limitation is primarily due to the persistent need for manual correction of speech transcriptions. 
Since diagnostic applications demand highly precise speech transcription and linguistic analysis, there is strong motivation to further enhance NLP approaches not based on commercial LLMs, such as by fine-tuning existing BERT-based POS tagging models with additional data and developing more advanced ASR models for Swiss German speech.

In Appendix~\ref{section:software}, we present our prototype software developed specifically for speech-language pathologists. As we continue to gather user feedback from real-world practice, we adhere to human-in-the-loop design principles and plan to enhance the software with more rigorously validated features for better user experience.

\section{Conclusion and Future Work}

In this study, we have addressed the challenges of automating key steps in language sample analysis by employing non-commercial NLP models for ASR and POS tagging. We evaluated the zero-shot capabilities of these models on both tasks and reported the empirical performance. Our work underscores the feasibility of employing ethical NLP approaches not based on commercial LLMs in the setting of speech-language pathology, particularly when handling sensitive data such as children's speech. In future work, we aim to expand the dataset by incorporating more diverse samples of children’s speech with and without DLD. Additionally, we plan to develop specialized ASR and POS tagging models tailored to Swiss German of children and evaluate whether analyses based on automatically created transcripts and automated POS tagging can reliably predict DLD in Swiss German-speaking children. We also plan to expand the linguistic analyses in broader contexts, especially on the syntactic level (such as dependency parsing, see the column \textbf{dependency} in Table~\ref{tab:annotation-examples} in Appendix~\ref{appendix:annotation-examples}), ultimately facilitating the semi-automatic diagnosis of DLD for children in Switzerland.

\section{Limitations}

The primary limitations of our work are as follows: (1) Due to the data sparsity and difficulty of data acquisition, our sample size is relatively low compared to evaluations in contexts other than the speech and language disorder context; (2) We did not conduct further investigations into morphological features, as, to the best of our knowledge, no existing NLP approaches not based on commercial LLMs for morphological prediction in Swiss German are currently available; (3) We did not benchmark the evaluation with locally deployed LLMs of the newest generation as this will be part of a future study; (4) All Swiss German ASR models known to us transcribe the text directly into Standard German. However, for our application, it would be beneficial for the text to be in Swiss German.  There is not yet any research showing that DLD approaches for other languages can be effectively applied to diagnosing DLD of Swiss German-speaking children. For this reason, the usefulness of the Standard German transcriptions obtained is limited; (5) Automatic tools for diagnosing DLD in children are not infallible; they should only be used in combination with other screening methods and by speech-language pathologists.

\section{Ethical Statement}

Our study received ethical approval from the responsible university's research committee. Informed consent was obtained from the parents of all participating children and the speech-language pathologists, allowing for the recording, processing, storage, and controlled sharing of data. To ensure accessibility, the consent form was provided in simplified language to facilitate understanding for parents. Children were informed about the study and provided their oral consent to participate.
None of the collected data have been processed through any commercial LLMs. All data processing was conducted locally. Permission was granted to publish the data from 41 recordings in a public repository. In addition to the small subset used for this paper, the whole dataset will be released in the near future.

\section{Acknowledgement}
This work was supported by the Foundation Special Education Centre Fribourg (Stiftung Heilpädagogisches Zentrum Fribourg) and the Foundation for Speech Therapy in the Canton of Zurich (Förderstiftung für das Sprachheilwesen im Kanton Zürich). We thank Susanne Kempe-Preti, Pascale Schaller, Julia Winkes, Sonja Schäli, Lena Graf, Ramona Rüegg, Lukas Fischer, Anne Göhring, Michelle Wastl and Angela Heldstab for their valuable input to the study. 

\section{Bibliographical References}
\label{sec:reference}

\bibliographystyle{lrec2026-natbib}
\bibliography{lrec2026-example}


\appendix
\onecolumn

\setcounter{footnote}{0}





\section{Instructions for Human Annotation of Part-of-speech Tagging on Gold Transcriptions (translated from the original German file)}

\paragraph{Task Description.}
\label{section:instruction}
Your task is to annotate a small dataset of spontaneous speech sentences of children with typical and atypical language development in Swiss German and Swiss Standard German. 
You received an Excel sheet with around 100 sentences from different children (K as \textit{Kind} (child) in German) and professionals (FP as \textit{Fachperson} (specialist) in German) interviewing the children in either Swiss German or Swiss Standard German. These sentences are randomly selected from different gold transcription files. In the first column you find the human transcriptions, while all other columns are generated automatically and need to be corrected. Please correct the annotations of: 

\begin{itemize}
    \item Part-of-speech tagging with the UPOS tags (more information here\footnote{\url{https://universaldependencies.org/u/pos/}}).
    \item Part-of-speech tagging with the STTS tags (more information here\footnote{\url{https://homepage.ruhr-uni-bochum.de/stephen.berman/Korpuslinguistik/Tagsets-STTS.html}}).
    \item Morphology for certain part-of-speech tags.
    \item Subject-verb agreement (SVA).
\end{itemize}

\paragraph{Annotation Process.} Please adhere to the following steps: 

\begin{enumerate}
\item Read this manual carefully.
\item Read the documentation of the tag sets.
\item Perform trial annotations using the control sentences constructed individually and send them back.
\item If you pass the trial annotations, you will receive real annotation sentences.
\item During the annotation, please keep in mind that:
\begin{itemize}
\item For the annotation you can use whatever assistance you want (such as look up tables, Duden, the internet, etc.);
\item Please \textbf{do not use the \texttt{spaCy} model \texttt{de\_core\_news\_sm} as well as the\texttt{
swiss\_german\_stts\_pos\_model} and the \texttt{swiss\_german\_pos\_model} models from Huggingface}, as they are baselines of our study;
\item Please \textbf{do not copy and paste the sentences into commercial LLMs}, as we must respect the data protection policy of Switzerland;
\item The sentences are random samples of different transcriptions. Please use only the current sentence as context;
\item You can annotate in whatever order you like.
\end{itemize}
\item For morphology, annotate as much as you can as long as it is determinable. If something is not determinable, please leave it out blank.
\item For subject-verb agreement:
\begin{itemize}
    \item Please mark all conjugated verbs with `v';
    \item Please mark the main part that determines the verb form (subjects) as `sb';
    \item For the contracted forms of Swiss German (such as ``gehen wir'' $\rightarrow$ ``g\"ommer'', ``kann er sie entsorgen'' $\rightarrow$ ``chanerse entsorge''), use sb\_v or v\_sb, depending on the order in the contraction.
\end{itemize}
\item Leave \texttt{<sentence>}, the sentence separator, as empty.
\item Tag all \texttt{UNK} as X/XY.
\item Tag all \texttt{NAME} (anonymized name) as PROPN/NE. 
\item Use the tag PROAV instead of PAV (adhering to \texttt{spaCy}).
\item In STTS, the verb \texttt{sein} is always tagged as VAxxx, even when used as a full verb. However, UPOS distinguishes between the functions of the verb: when a verb is used as a full verb, can can thus be replaced by \texttt{sich befinden}, it is tagged as VERB.
\item The STTS tag set labels all occurrences of auxiliary verbs (\texttt{sein, werden, haben}) as well as copula with VAxxx, modal verbs (\texttt{m\"ussen, d\"urfen, wollen, m\"ogen, k\"onnen, sollen}) are labelled with VMxxx.
\item Ja/Nein: Please annotate as PTKANT if
\begin{itemize}
    \item Standing alone;
    \item Is part of an answer;
    \item Used as a query (R\"uckfrage).
\end{itemize}
Except when used as modal particle (in German ``Abt\"onungspartikel'', for instance, ``Das ist ja wirklich sch\"on''), annotate as ADV. 
\item Interjections encompass:
\begin{itemize}
    \item All signals of understanding (in German ``Verst\"andigungssignale'', for instances, ``oh', ``ah'', ``aha'', ``wow'', ``hmm'');
    \item All onomatopoeias (for instances, ``brummbrumm'', ``gluglug'', ``miau'', ``wuff'').
\end{itemize}
\item Corrections, word fragments, errors (which are then corrected, for instance, ``eine Bri-Brille''), please annotate as X/XY.
\item If a child made a grammatical error and used the wrong form, \textbf{please annotate what was exactly said and not what it should be}, as we want to identify these errors later. If not all information can be clearly determined from the used word itself, assume the correct form.
\item For cases that are hard to tag (e.g., the child used irregular forms, wrong or fantasy words, etc.), please do your best.
\end{enumerate}

\paragraph{Rules of Swiss German.} Most rules are identical between Swiss German and Standard German. However, there are still some linguistic differences. Please read the chapter \textit{Differences to German UD Guidelines} here\footnote{\url{https://universaldependencies.org/gsw/}} as a reference.

\paragraph{Conversion between UPOS tag set and STTS tag set.} 
See more information here\footnote{\url{https://universaldependencies.org/tagset-conversion/de-stts-uposf.html}}. Please be aware that the conversion does not apply to all cases.

\newpage
\section{Canonical Examples of POS Tagging Disagreements between Human Annotators}
\label{section:pos-tag-error}

\subsection{Examples in Swiss German}
\paragraph{Disagreement 1} Whether a word is a proper noun (PROPN), noun (NOUN), or foreign word (X).
\begin{table*}[htb]
    \centering
    \begin{tabular}{lcccccccccc}
    \toprule
     & ebe & genau & ja & de & het & der & paw & patrol & gfalle & ?  \\
     A & ADV & ADV & PART & ADV & AUX & PRON & \textcolor{BrickRed}{\textit{X}} & \textcolor{BrickRed}{\textit{X}} & VERB & PUNCT \\
     B & ADV & ADV & PART & ADV & AUX & PRON & \textcolor{BrickRed}{\textit{PROPN}} & \textcolor{BrickRed}{\textit{PROPN}} & VERB & PUNCT \\
     C & ADV & ADV & PART & ADV & AUX & PRON & \textcolor{BrickRed}{\textit{PROPN}} & \textcolor{BrickRed}{\textit{PROPN}} & VERB & PUNCT \\
     \bottomrule
    \end{tabular}
    \caption{Example sentence corresponding to the English translation \textit{so exactly then you liked paw patrol?}}
\end{table*}
\paragraph{Disagreement 2} Whether a word is an adverb (ADV) or, e.g., an adjective (ADJD) or interjection (ITJ).
\begin{table*}[!htb]
    \centering
    \begin{tabular}{lccccc}
    \toprule
     & ja & hm & genau & ah & schpannend  \\
     A & PTKANT & ITJ & \textcolor{BrickRed}{\textit{ADJD}} & ITJ & ADJD \\
     B & PTKANT & ITJ & \textcolor{BrickRed}{\textit{ADV}} & ITJ & ADJD \\
     C & PTKANT & ITJ & \textcolor{BrickRed}{\textit{ADV}} & ITJ & ADJD \\
     \bottomrule
    \end{tabular}
    \caption{Example sentence corresponding to the English translation \textit{yes hm exactly ah interesting}.}
\end{table*}
\begin{table*}[!htb]
    \centering
    \begin{tabular}{lcccccc}
    \toprule
     & g\"all & was & d\"anksch & was & isch & das \\
     A & \textcolor{BrickRed}{\textit{ITJ}} & PWS & VVFIN+ & PWS & VAFIN & PDS \\
     B & \textcolor{BrickRed}{\textit{ADV}} & PWS & VVFIN & PWS & VAFIN & PDS \\
     C & \textcolor{BrickRed}{\textit{ITJ}} & PWS & VVFIN+ & PWS & VAFIN & PDS \\
     \bottomrule
    \end{tabular}
     \caption{Example sentence corresponding to the English translation \textit{what do you think what this is?}}
\end{table*}

\paragraph{Disagreement 3} Distinguishing different types of pronouns (e.g., PDS, PDAT).
\begin{table*}[!htb]
    \centering
    \begin{tabular}{lcccccc}
    \toprule
    & gend & recht & gas & uf & dene & trotti \\
    A & VVFIN & ADV & NN & APPR & \textcolor{BrickRed}{\textit{PDS}} & NN  \\
    B & VVFIN & ADV & PTKVZ & APPR & \textcolor{BrickRed}{\textit{PDAT}} & NN \\
    C & VVFIN & ADV & NN & APPR & \textcolor{BrickRed}{\textit{PDS}} & NN \\
    \bottomrule
    \end{tabular}
     \caption{Example sentence corresponding to the English translation \textit{they really step on the gas on these scooters}}
\end{table*}

\paragraph{Disagreement 4} Whether to tag attributive pronouns (PPOSAT in STTS)  as determiner (DET) or pronoun (PRON) in UPOS.
\begin{table*}[!htb]
    \centering
    \begin{tabular}{lccccccccc}
    \toprule
     & oh & verzell & mal & wy & fyrsch & dù\footnotemark & din & gebùrtstag & ?  \\
     
     A & INTJ & VERB & ADV & ADV & VERB & PRON & \textcolor{BrickRed}{\textit{DET}} & NOUN & PUNCT \\
     B & INTJ & VERB & ADV & CCONJ & VERB & PRON & \textcolor{BrickRed}{\textit{PRON}} & NOUN & PUNCT \\
     C & INTJ & VERB & ADV & ADV & VERB & PRON & \textcolor{BrickRed}{\textit{DET}} & NOUN & PUNCT \\
     STTS & ITJ & VVIMP & ADV & PWAV & VVFIN & PPER & \textcolor{BrickRed}{\textit{PPOSAT}} & NN & \$. \\
     \bottomrule
    \end{tabular}
     \caption{Example sentence corresponding to the English translation \textit{oh tell me how do you celebrate your birthday?}}
\end{table*}
\footnotetext{Adding an accent symbol to a vowel describes its quality adapted from \citet{dieth1986schwyzertütsch}.}

\paragraph{Disagreement 5} Whether a word is a concatenation of multiple words (e.g., VVFIN+) or not (e.g., VVFIN) (especially in the case of a verb in second person singular, as it can stand without PPER in Swiss German).
\begin{table*}[!htb]
    \centering
    \begin{tabular}{lcccccc}
    \toprule
     & g\"all & was & d\"anksch & was & isch & das \\
     A & ITJ & PWS & \textcolor{BrickRed}{\textit{VVFIN+}} & PWS & VAFIN & PDS \\
     B & ADV & PWS & \textcolor{BrickRed}{\textit{VVFIN}} & PWS & VAFIN & PDS \\
     C & ITJ & PWS & \textcolor{BrickRed}{\textit{VVFIN+}} & PWS & VAFIN & PDS \\
     \bottomrule
    \end{tabular}
     \caption{Example sentence corresponding to the English translation \textit{what do you think this is?}}
\end{table*}

\subsection{Examples in Swiss Standard German}

\paragraph{Disagreement 1} Whether a word is nominalized (used as a noun) or not.
\begin{table*}[!htb]
    \centering
    \resizebox{\textwidth}{!}{
     \begin{tabular}{lcccccccccccc}
    \toprule
    & aber & du & hast & recht & das & macht & man & doch & eigentlich & mit & einem & stock \\
    A & CCONJ & PRON & AUX & \textcolor{BrickRed}{\textit{NOUN}} & PRON & VERB & PRON & ADV & ADV & ADP & DET & NOUN \\
    B & CCONJ & PRON & AUX & \textcolor{BrickRed}{\textit{NOUN}} & PRON & VERB & PRON & ADV & ADV & ADP & DET & NOUN \\
    C & CCONJ & PRON & AUX & \textcolor{BrickRed}{\textit{ADV}} & PRON & VERB & PRON & ADV & ADV & ADP & DET & NOUN \\
    \bottomrule
    \end{tabular}
    }
    \caption{Example sentence corresponding to the English translation \textit{but you are right actually you do this with a stick}.}
\end{table*}

\paragraph{Disagreement 2} For incomplete or erroneous words, what is the right way to interpret it (the correct word in the first example should be ``meinte'' (``meant'') as one word, and in the second example, ``warte'' (``wait'').

\begin{table*}[!htb]
    \centering
    \begin{subtable}[t]{0.45\textwidth}
        \centering
        \begin{tabular}{lccc}
        \toprule
         & mein & te & an  \\
         A & \textcolor{BrickRed}{\textit{PPOSAT}} & \textcolor{BrickRed}{\textit{NN}} & PTKVZ \\
         B & \textcolor{BrickRed}{\textit{PPOSAT}} & \textcolor{BrickRed}{\textit{XY}} & XY \\
         C & \textcolor{BrickRed}{\textit{VVFIN}} & \textcolor{BrickRed}{\textit{VVFIN}} & APPR \\
         \bottomrule
        \end{tabular}
        \caption{English translation of the example sentence: \textit{mean-ed on}.}
    \end{subtable}
    \hspace{1em}
    \begin{subtable}[t]{0.45\textwidth}
        \centering
        \begin{tabular}{lccc}
        \toprule
         & wate & an & freitag  \\
         A & \textcolor{BrickRed}{\textit{NN}} & APPR & NN \\
         B & \textcolor{BrickRed}{\textit{XY}} & APPR & NN \\
         C & \textcolor{BrickRed}{\textit{VVIMP}} & APPR & NN \\
         \bottomrule
        \end{tabular}
        \caption{Example sentence corresponding to the English translation \textit{wait on friday}.}
    \end{subtable}
    \caption{Comparative morphological annotation examples from two German utterances.}
\end{table*}

\paragraph{Disagreement 3} Whether a verb is used as auxiliary (VA) or full verb (VV) (differences exist between the two tag sets).
\begin{table*}[!htb]
    \centering
    \resizebox{\textwidth}{!}{
    \begin{tabular}{lcccccccccccc}
    \toprule
     & ja & diese & zeitung & ich & hab & das & gerne & aber & ich & hat & ein & zeichnung  \\
     A (UPOS) & PART & DET & NOUN & PRON & VERB & PRON & ADV & CCONJ & PRON & \textcolor{BrickRed}{\textit{AUX}} & DET & NOUN \\
     A (STTS) & PTKANT & PDAT & NN & PPER & VAFIN & PDS & ADV & KON & PPER & \textcolor{BrickRed}{\textit{VAFIN}} & ART & NN \\
     B (UPOS) & PART & DET & NOUN & PRON & VERB & PRON & ADV & CCONJ & PRON & \textcolor{BrickRed}{\textit{AUX}} & DET & NOUN \\
     B (STTS) & PTKANT & PDAT & NN & PPER & VAFIN & PDS & ADJD & KON & PPER & \textcolor{BrickRed}{\textit{VAFIN}} & ART & NN \\
     C (UPOS) & PART & DET & NOUN & PRON & VERB & PRON & ADV & CCONJ & PRON & \textcolor{BrickRed}{\textit{VERB}} & DET & NOUN \\
     C (STTS) & PTKANT & PDAT & NN & PPER & VAFIN & PDS & ADV & KON & PPER & \textcolor{BrickRed}{\textit{VAFIN}} & ART & NN \\
     \bottomrule
    \end{tabular}
    }
    \caption{Example sentence corresponding to the English translation \textit{yes this newspaper I like this but I has a drawing}.}
\end{table*}

\newpage
\section{Per-speaker-group ASR and POS Tagging Results}

\paragraph{Observation 1} Whisper has significantly less ASR errors on transcriptions of speech-language pathologists than of children.
\begin{table*}[!htb]
    \centering
    \begin{tabular}{lcccccccccccc}
    \toprule
     \multirow{2}{*}{\textbf{Original Transcripts}} & \multicolumn{2}{c}{\textbf{WER} $\downarrow$} & & \multicolumn{2}{c}{\textbf{CER} $\downarrow$} & & \multicolumn{2}{c}{\textbf{MER} $\downarrow$} & & \multicolumn{2}{c}{\textbf{WIL} $\downarrow$} &   \\
     \cmidrule{2-3}
     \cmidrule{5-6}
     \cmidrule{8-9}
     \cmidrule{11-12}
      & FP & K & & FP & K & & FP & K & & FP & K \\
     \hline
     Swiss German & 0.772 & 0.860 & & 0.469 & 0.556 & & 0.761 & 0.850 & & 0.927 & 0.970 \\ 
     Swiss Std. German & 0.377 & 0.600 & & 0.278 & 0.445 & & 0.370 & 0.591 & & 0.463 & 0.722 \\
     \hline
    \end{tabular}
    \caption{ASR results evaluated on \textbf{original} speech transcriptions of speech-language pathologists (FP) and children (K).}
\end{table*}
\begin{table*}[!htb]
    \centering
    \begin{tabular}{lcccccccccccc}
    \toprule
     \multirow{2}{*}{\textbf{Normalized Transcripts}} & \multicolumn{2}{c}{\textbf{WER} $\downarrow$} & & \multicolumn{2}{c}{\textbf{CER} $\downarrow$} & & \multicolumn{2}{c}{\textbf{MER} $\downarrow$} & & \multicolumn{2}{c}{\textbf{WIL} $\downarrow$} &   \\
     \cmidrule{2-3}
     \cmidrule{5-6}
     \cmidrule{8-9}
     \cmidrule{11-12}
      & FP & K & & FP & K & & FP & K & & FP & K \\
     \hline
     Swiss German & 0.506 & 0.681 & & 0.337 & 0.466 & & 0.494 & 0.664 & & 0.666 & 0.840 \\ 
     Swiss Std. German & 0.365 & 0.550 & & 0.275 & 0.430 & & 0.359 & 0.541 & & 0.444 & 0.651 \\
     \hline
    \end{tabular}
    \caption{ASR results evaluated on \textbf{normalized} speech transcriptions of speech-language pathologists (FP) and children (K).}
\end{table*}
\paragraph{Observation 2} POS tagging models
have higher F1 scores on transcriptions of speech-language pathologists (i.e. adults) than of children.
\begin{table*}[!htb]
    \centering
    \begin{tabular}{lcccccc}
    \toprule
    \multirow{2}{*}{\textbf{Original Transcriptions}} & \multicolumn{2}{c}{\textbf{UPOS} (\textbf{F1 Score} $\uparrow$)} & & \multicolumn{2}{c}{\textbf{STTS} (\textbf{F1 Score} $\uparrow$)} \\ 
    \cmidrule{2-3}
    \cmidrule{5-6}
    & FP & K & & FP & K \\
    \hline 
    Swiss German & 0.735 & 0.673 & & 0.728 & 0.675 \\
    Swiss Std. German & 0.844 & 0.659 & & 0.831 & 0.654 \\
    \bottomrule
    \end{tabular}
    \caption{POS tagging results on \textbf{original} speech transcriptions of speech-language pathologists (FP) and children (K).}
\end{table*}
\paragraph{Observation 3} Swiss Standard German transcriptions achieved generally higher evaluation results than Swiss German transcriptions. 
\begin{table*}[!htb]
    \centering
    \begin{tabular}{lcccccc}
    \toprule
    \multirow{2}{*}{\textbf{Normalized Transcriptions}} & \multicolumn{2}{c}{\textbf{UPOS} (\textbf{F1 Score} $\uparrow$)} & & \multicolumn{2}{c}{\textbf{STTS} (\textbf{F1 Score} $\uparrow$)} \\ 
    \cmidrule{2-3}
    \cmidrule{5-6}
    & FP & K & & FP & K \\
    \hline 
    Swiss German & 0.835 & 0.787 & & 0.823 & 0.762  \\
    Swiss Std. German & 0.878 & 0.796 & & 0.852 & 0.807 \\
    \bottomrule
    \end{tabular}
    \caption{POS tagging results on \textbf{normalized} speech transcriptions of speech-language pathologists (FP) and children (K).}
\end{table*}
\paragraph{Observation 4} Orthographic normalization of speech transcriptions helps in boosting the performance of both ASR model and POS tagging model. 

\newpage
\section{The LSA Software for Therapeutic Practice of DLD Diagnosis}
\label{section:software}

\begin{figure*}[!htb]
    \centering
    \includegraphics[width=0.7\textwidth]{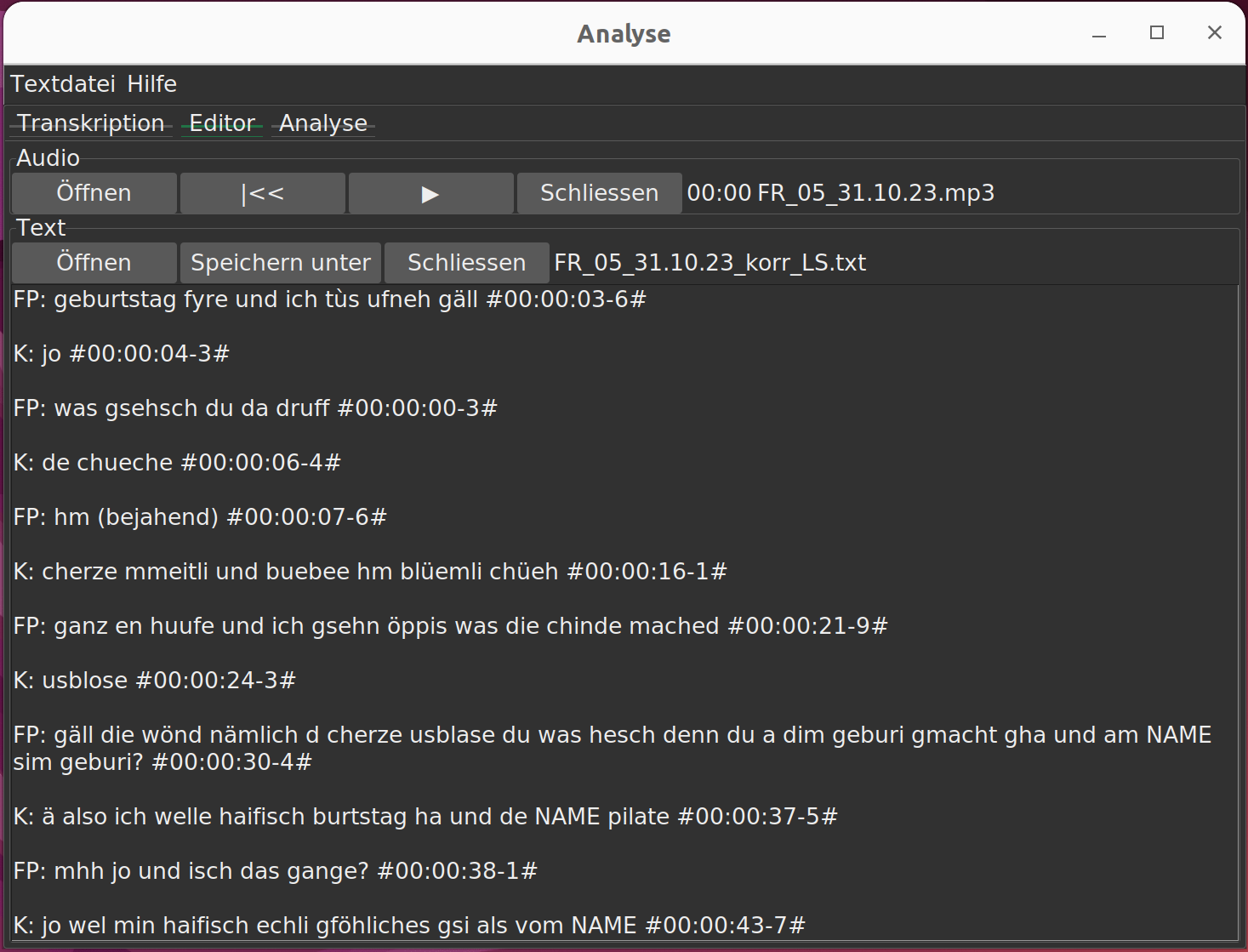}
    \caption{Editor of the software. After automatically transcribing the recordings, the transcript is opened in the editor. Here, the recordings can be played and the transcript can be corrected.}
\end{figure*}
\begin{figure*}[!htb]
    \centering
    \includegraphics[width=0.7\textwidth]{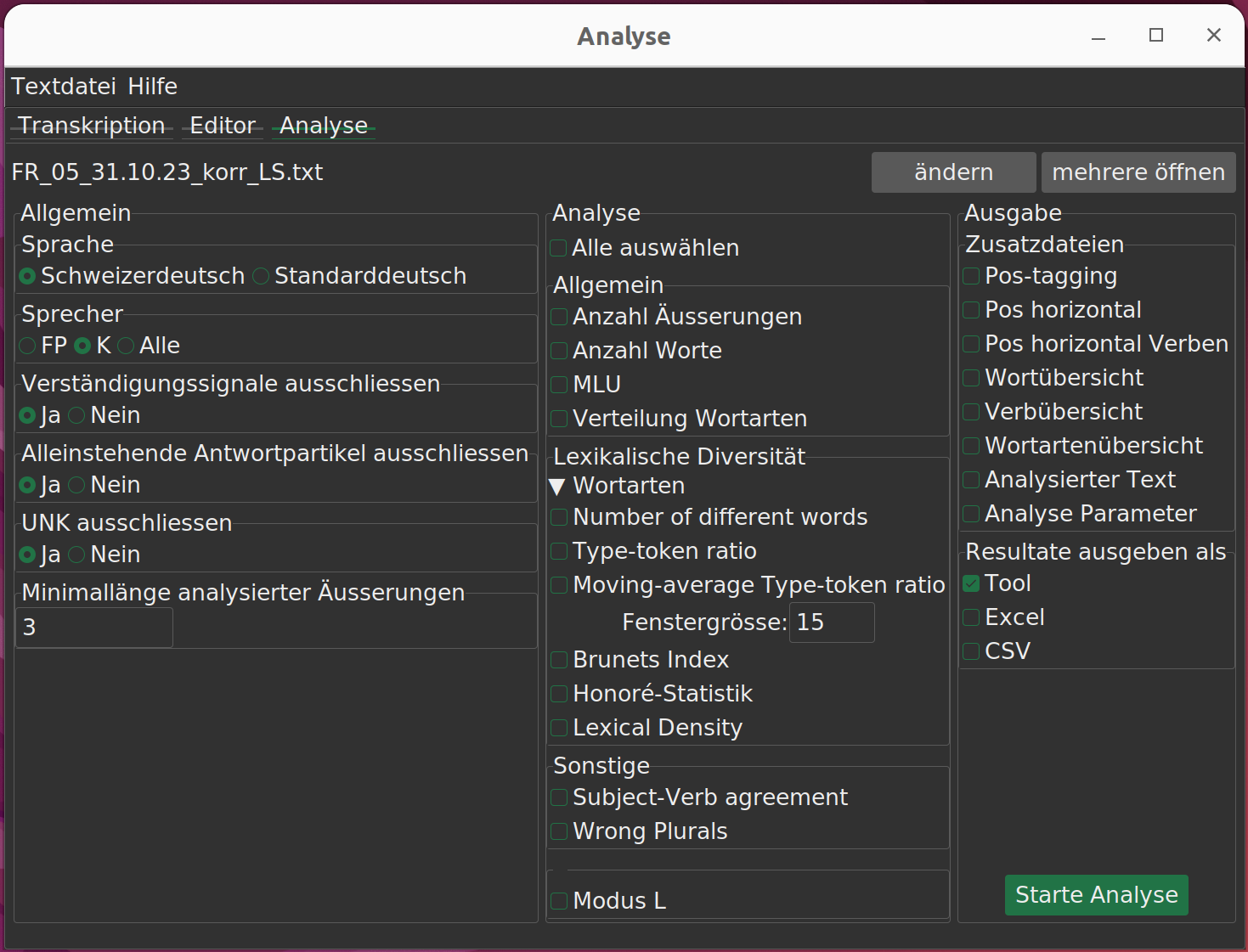}
    \caption{Analysis: After correcting the automatic transcription manually, the analysis can be started. Providing different options for the analysis, such as which speakers should be analyzed and what should be filtered out, a personalized analysis can be executed, containing values such as mean length of utterance, distribution of POS tags, and subject verb agreement as well as additional files with overviews of all verbs and POS tagging.}
\end{figure*}

\vspace{5em}

\begin{sidewaystable*}[]
  \centering

  \section{Annotation Examples} \label{appendix:annotation-examples}

  \vspace{3em}
  
  \begin{adjustbox}{max width=\textwidth, max height=\textheight} 

   
    \begin{tabular}{cllllllllllll}
    \toprule
    \textbf{send\_id} & \textbf{speaker} & \textbf{word\_id} & \textbf{word} & \textbf{normalized} & \textbf{lemma} & \textbf{UPOS tag} & \textbf{STTS tag} & \textbf{morphology} & \textbf{SVA} & \textbf{dependency} \\
    \midrule
    62 & FP & 0 & warst & Warst & sein & VERB & VAFIN & \{`Mood': `Ind', `Number': `Sing', `Person': `2', `Tense': `Past', `VerForm': `Fin'\} & v & ROOT \\
    62 & FP & 1 & du & du & du & PRON & PPER & \{`Case': `Norm', `Number': `Sing', `Person': `2', `PronType': `Prs'\} & sb & sb \\
    62 & FP & 2 & diesen & diesen & dieser & DET & PDAT & \{`Case': `Acc', `Gender': `Masc', `Number': `Sing', `PronType': `Dem'\} & & nk \\
    62 & FP & 3 & sommer & Sommer & Sommer & NOUN & NN & \{`Case': `Acc', `Gender': `Masc', `Number': `Sing'\} & & oa \\
    62 & FP & 4 & auch & auch & auch & ADV & ADV & & & mo \\
    62 & FP & 5 & schon & schon & schon & ADV & ADV & & & mo \\
    62 & FP & 6 & in & in & in & ADP & APPR & & & mo \\
    62 & FP & 7 & der & der & der & DET & ART & \{`Case': `Dat', `Definite': `Def', `Gender': `Fem', `Number': `Sing', `PronType': `Art'\} & & nk \\
    62 & FP & 8 & badi & Badi & Badi & NOUN & NN & \{`Case': `Dat', `Gender': `Fem', `Number': `Sing'\} & & oa \\
    62 & FP & 9 & ? & ? & ? & PUNCT & \$. & & & punct \\
    63 & K & 0 & \"ahm & \"Ahm & \"ahm & INTJ & ITJ & & & mo \\
    63 & K & 1 & ja & ja & ja & PART & PTKANT & & & mo \\
    63 & K & 2 & abeR & aber & aber & CCONJ & KON & & & mo \\
    63 & K & 3 & is & ist & sein & AUX & VAFIN & \{`Mood': `Ind', `Number': `Sing', `Person': `3', `Tense': `Pres', `VerbForm': `Fin'\} & v & sb \\
    63 & K & 4 & de & der & der & PRON & PDS & \{`Case': `Nom', `Gender': `Masc', `Number': `Sing', `PronType': `Dem'\} & sb & uc \\
    63 & K & 5 & mine & meine & mein & PRON & PPOSAT &  \{`Case': `Nom', `Gender': `Neut', `Number': `Sing', `Poss': `Yes', `PronType': `Prs'\} & & nk \\ 
    63 & K & 6 & haus & Haus & Haus & NOUN & NN & \{`Case': `Nom', `Gender': `Neut', `Number': `Sing'\} & & ams \\
    63 & K & 7 & z\"un & z\"un & z\"un & X & XY & & & pd \\
    63 & K & 8 & ist & ist & sein & VERB & VAFIN & \{`Mood': `Ind', `Number': `Sing', `Person': `3', `Tense': `Pres', `VerbForm': `Fin'\} & v & ROOT \\
    63 & K & 9 & da & da & da & ADV & ADV & & & mo \\
    63 & K & 10 & drauf & drauf & drauf & ADV & PROAV & & & mo \\
    63 & K & 11 & badi & Badi & Badi & NOUN & NN & \{`Case': `Nom', `Gender': `Fem', `Number': `Sing'\} & sb & oc \\
    63 & K & 12 & kann & kann & k\"onnen & AUX & VMFIN & \{`Mood': `Ind', `Number': `Sing', `Person': `3', `Tense': `Pres', `VerbForm': `Fin'\} & v & sb \\
    63 & K & 13 & kann & kann & k\"onnen & AUX & VMFIN & \{`Mood': `Ind', `Number': `Sing', `Person': `3', `Tense': `Pres', `VerbForm': `Fin'\} & v & sb \\
    63 & K & 14 & t\"ure & T\"ure & T\"ur & NOUN & NN & \{`Case': `Acc', `Gender': `Fem', `Number': `Sing'\} & & sb \\
    63 & K & 15 & offen & offen & offen & ADV & ADJD & \{`Degree': `Pos'\} & & oc \\
    63 & K & 16 & und & und & und & CCONJ & KON & & & cd \\
    63 & K & 17 & da & da & da & ADV & ADV & & & mo \\
    63 & K & 18 & ist & ist & sein & VERB & VAFIN & \{`Mood': `Ind', `Number': `Sing', `Person': `3', `Tense': `Pres', `VerbForm': `'\} & v & cj \\
    63 & K & 19 & so & so & so & ADV & ADV & & & mo \\
    63 & K & 20 & bile & viele & vieler & PRON & PIAT & \{`Case': `Nom', `Gender': `Fem', `Number': `Plur', `PronType': `Ind'\} & & sb \\
    63 & K & 21 & badi & Badi & Badi & NOUN & NN & \{`Case': `Nom', `Gender': `Fem', `Number': `Sing'\} & sb & pd \\
    63 & K & 22 & und & und & und & CCONJ & KON & & & cd \\
    63 & K & 23 & kann & kann & k\"onnen & AUX & VMFIN & \{`Mood': `Ind', `Number': `Sing', `Person': `3', `Tense': `Pres', `VerbForm': `Fin'\} & v & cj \\
    63 & K & 24 & dach & Dach & Dach & NOUN & NN & \{`Case': `Nom', `Gender': `Neut', `Number': `Sing'\} & & mo \\
    63 & K & 25 & da & da & da & ADV & ADV & & & mnr \\
    63 & K & 26 & badi & Badi & Badi & NOUN & NN & \{`Case': `Nom', `Gender': `Fem', `Number': `Sing'\} & & mo \\
    63 & K & 27 & lauffen & laufen & laufen & VERB & VVINF & \{`VerbForm': `Inf'\} & & oc \\
    \bottomrule
    \end{tabular}
  \end{adjustbox}
  \caption{Swiss Standard German language sample annotated. Words like ``abeR'' with R in upper case indicate the emphasizing tone.}
  \label{tab:annotation-examples}
\end{sidewaystable*}

\end{document}